\documentclass[11pt,a4paper]{article}
\usepackage[hyperref]{naacl2021}
\usepackage{times}
\usepackage{latexsym}

\usepackage{graphicx}
\usepackage{amsmath}
\usepackage{makecell}
\usepackage{booktabs}
\usepackage{multirow}
\usepackage[normalem]{ulem}
\usepackage{cancel}
\usepackage{cleveref}
\crefname{section}{§}{§§}
\Crefname{section}{§}{§§}
\usepackage[utf8]{inputenc}
\usepackage{enumitem}
\usepackage{tikz}
\usetikzlibrary{tikzmark}
\usepackage{hyperref}
\usepackage[none]{hyphenat}

\usepackage{url}

\crefname{section}{§}{§§}
\Crefname{section}{§}{§§}

\usepackage{microtype}

\title{CodemixedNLP: An Extensible and Open NLP Toolkit for Code-Mixing}

\author{
Sai Muralidhar Jayanthi,
Kavya Nerella,
Khyathi Raghavi Chandu, 
Alan W Black \\
Language Technologies Institute \\
Carnegie Mellon University \\
\texttt{\{sjayanth, knerella, kchandu, awb\}@cs.cmu.edu}
}

\date{}

\begin{document}
\maketitle

\begin{abstract}
The NLP community has witnessed steep progress in a variety of tasks across the realms of monolingual and multilingual language processing recently. These successes, in conjunction with the proliferating mixed language interactions on social media have boosted interest in modeling code-mixed texts. In this work, we present \textsc{CodemixedNLP}, an open-source library with the goals of bringing together the advances in code-mixed NLP and opening it up to a wider machine learning community. The library consists of tools to develop and benchmark versatile model architectures that are tailored for mixed texts, methods to expand training sets, techniques to quantify mixing styles, and fine-tuned state-of-the-art models for 7 tasks in Hinglish.
We believe this work has a potential to foster a distributed yet collaborative and sustainable ecosystem in an otherwise dispersed space of code-mixing research.
The toolkit is designed to be simple, easily extensible, and resourceful to both researchers as well as practitioners\footnote{The library \& pretrained models are available at \url{https://github.com/murali1996/CodemixedNLP}.}.
\end{abstract}

\section{Introduction}
\label{sec:introduction}


Code-mixing
refers to fluid alteration between two or more languages in a given
utterance. 
This phenomenon is ubiquitous and more natural 
in multilingual communities, and is highly prevalent in social media platforms.
Developing tools that can comprehend mixed texts can have a 
multitude of advantages, ranging from socially responsible NLP applications such as 
moderating abusive content in social media to 
improve naturalness of ubiquitous technologies such as  conversational AI assistants and further to 
develop socio-cultural studies around human cognition, such as
why and when people code-mix.

NLP 
tools
for
monolingual and multilingual language processing have 
rapidly progressed in the past few years; 
thanks to the transformer-based models such as
Multilingual BERT \cite{devlin-etal-2019-bert} 
\& XLM-RoBERTa \cite{conneau-etal-2020-unsupervised},
and their \textit{pretraining} techniques.
On various mixed datasets, recent studies have shown that
adopting 
multilingual pretrained models
can perform better than their previous deep learning counterparts
\cite{pires-etal-2019-multilingual, khanuja-etal-2020-gluecos, aguilar-etal-2020-lince, chakravarthy-etal-2020-detecting, jayanthi2021sjajdravidianlangtecheacl2021}.
While this looks promising for multilingual, the same is not translated to code-mixing.
Hence,
a critical investigation is required to understand
generalizable modeling
strategies to  
enhance performance on mixed texts
\cite{winata2021multilingual, aguilar-solorio-2020-english, sitaram2020survey}.

At the same time, 
practitioners who 
require
an off-the-shelf tool 
into their downstream mixed text application
(eg. sentiment or language identification), 
currently have 
to resort to monolingual toolkits such as 
\citet{nltk}, 
\citet{flair}, 
\citet{indicnlp}
and
\citet{inltk}.
On the other hand, while there have been several episodic works on mixed text processing,
such as proposing novel datasets or shared-tasks or training strategies, there haven't been 
many initiatives to collate these
resources into a common setting; doing so can benefit both
researchers and practitioners, thereby accelerating NLP for mixed texts.

In this work, we address 
some of these shortcomings
by creating an 
extensible and open-source toolkit for a variety of semantic
and syntactic NLP applications in mixed languages. Our toolkit offers-
\begin{itemize}[noitemsep,topsep=0pt] 
    \item simple \textbf{plug-and-play command line interfaces} with fine grained control over inputs, models and tasks for developing, quantifying, benchmarking, and re-using versatile model architectures tailored for mixed texts 
    (\cref{ssec:inputs}, \cref{ssec:models}, \cref{ssec:tasks})
    \item easy to use single stop interfacing for a variety of \textbf{data augmentation} techniques including transliteration, spelling variations, expansion 
    with
    monolingual corpora etc., by leveraging a collation of publicly available tools
    \item a toolkit library to \textbf{import fine-tuned and ready-to-use models} for 7 different tasks in 
    Hinglish, along with an easy-to-setup \textbf{web interface} wrapper based on flask server (\cref{sec:demo})
\end{itemize}

We believe the fine grained plug and play interfacing of the  toolkit can serve a multitude of purposes in both academia and industry. Such fine control over the individual components of the model can enable accelerated experimentation in training different model architectures, 
such as multi-tasking, representation-fusion, and language-informed modeling.
This in-turn 
helps our understanding of utilizing pretrained transformer-models for
mixed datasets. In addition, our toolkit also offers computation of metrics to quantify code-mixing such as
Code-Mixing Index, Language Entropy, etc., which can be utilized to
find peculiarities of low-performing subsets.


Like a curse in disguise, though code-mixing is widely prevalent and available on social media, it is accompanied with 
non-standard spellings, mixed scripts and 
ill-formed sentences are common in code-mixing.
To combat this, our toolkit offers techniques to augment the training sets with multiple views of each input corresponding to the above problems. 

Among many potential applications, 
we first demonstrate our toolkit's utility 
in \textit{benchmarking} (\cref{sec:experiments}).
In addition, we publish 
state-of-the-art models
for  different NLP tasks in Hinglish 
and wrap them into a command line / deployable
web interface (\cref{sec:demo}).
Our toolkit is easily extensible-- practitioners can incorporate
new pretrained as well as fine-tuned models, 
include text processors such as tokenizers, transliterators and translators, 
and 
add wrappers on existing methods for downstream NLP applications.

\section{Toolkit}
\label{sec:toolkit}

\begin{figure}
\centering
\includegraphics[width=0.4\textwidth]{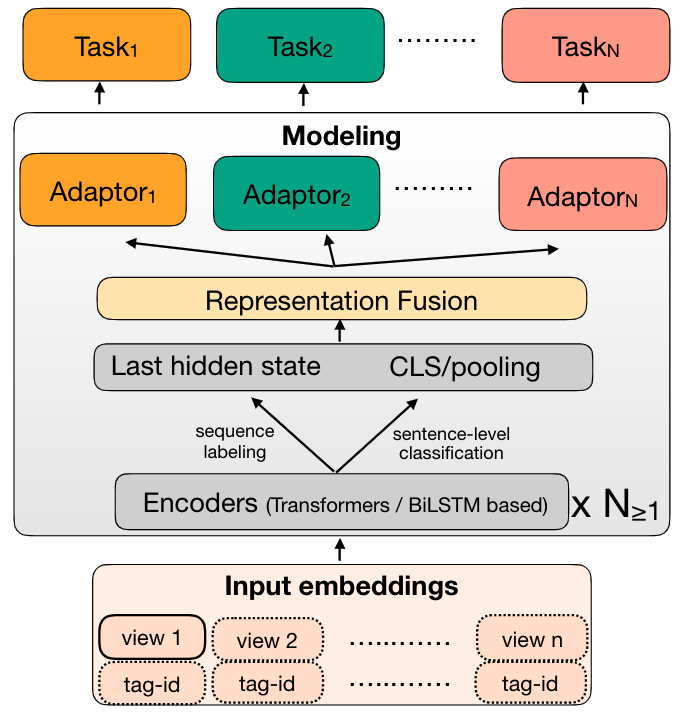}
\caption{\small Customizable components in our toolkit. Marked in dashed box is an optional component.}
\label{fig:components}
\end{figure}

Our toolkit is organized into components
as depicted in Figure~\ref{fig:components}. In a nutshell, 
an end-to-end model architecture consists
of one or more \texttt{encoder} components, 
a component for \texttt{combining encodings}, and one or more \texttt{adaptor} plus \texttt{task} components.

\begin{table*}[!ht]
\resizebox{1.0\textwidth}{!}{%
\begin{tabular}{l|ccccccc|ccc}
    
    \multicolumn{11}{c}{\textbf{Benchmarking transformer based models} (F1 / Accuracy)} \\ \hline 
    
    \hline 
    \multicolumn{1}{c|}{} & 
    \multicolumn{7}{c|}{Text Classification Tasks} &
    \multicolumn{3}{c}{Sequence Tagging Tasks} \\ 
    \cline{2-11}
    
    \multicolumn{1}{c|}{}
    & \multicolumn{3}{c|}{Sentiment  Classification}
    & \multicolumn{1}{c|}{\makecell{Aggression \\ Identification}}
    & \multicolumn{1}{c|}{\makecell{Hate Speech \\ Identification}}
    & \multicolumn{1}{c|}{\makecell{Offensiveness \\ Identification}}
    & \multicolumn{1}{c|}{\makecell{Youtube \\ Comments\\Classification}}
    & \multicolumn{1}{c|}{\makecell{Language \\ Identification}}
    & \multicolumn{1}{c|}{NER}
    & \multicolumn{1}{c}{POS}
    \\ 
    \cline{2-11}
    

    \multicolumn{1}{c|}{Model} 
    & \multicolumn{1}{c}{HIN-ENG$_{1}$}
    & \multicolumn{1}{c}{HIN-ENG$_{2}$}
    & \multicolumn{1}{c|}{SPA-ENG$^{\dagger}$}
    & \multicolumn{1}{c|}{HIN-ENG}
    & \multicolumn{1}{c}{HIN-ENG}
    & \multicolumn{1}{c|}{TAM-ENG$^{\dagger}$}
    & \multicolumn{1}{c|}{HIN-ENG}
    & \multicolumn{1}{c|}{HIN-ENG$^{\dagger}$}
    & \multicolumn{1}{c|}{HIN-ENG$^{\dagger}$}
    & \multicolumn{1}{c}{HIN-ENG} 
    \\
    
    \hline
    
    \texttt{mBert} & 65.9 / 65.7 & 58.8 / 60.7	& 50.1 / 51.3 &	48.1 \ 48.7 &  	47.1 / 61.7 & 76.8 / 79.0	 & 83.3 / 83.4 & 96.9 / 96.9 & 95.1 / 95.6 & 75.0 / 75.8\\
    \hspace{3mm} w/ Task adaptive & 67.4 / 67.2 & 61.4 / 61.5 &	53.2 / 55.0 & 50.3 / 51.2 & 67.9 \ 70.3 & 76.9 / \textbf{78.9} & 83.6 / 83.7	 & 97.1 / 97.1 & 97.1 / 97.1 &  \textbf{77.1 / 77.6}\\ 
    \hspace{3mm} w/ Domain adaptive & 71.4 / 71.3 & 62.5 / 63.0 & $-$ & 50.7 / 51.4 & \textbf{65.5 / 70.3} & $-$ & 85.2 / 85.3 &	97.3 / 97.3 & \textbf{97.2 / 97.3 } & 75.8 / 76.3 \\
    \hline
    
    \texttt{XLM-RoBERTa} & 68.9 / 69.1 &	61.5 / 61.5	& 54.4 / 54.6 & 49.0 / 49.6 & 64.4 / 69.6 & 76.7 / 77.4 & 85.7 / 85.8 &	97.1 / 97.1 & 96.1 / 96.3 &	 73.7 / 74.8 \\
    \hspace{3mm} w/ Task adaptive & 70.4 / 70.4 &	63.0 / 63.0	 & \textbf{54.4 / 54.8}	 & 55.3 / 55.4 & 64.6 / 69.8 & \textbf{77.1} / 78.8 & 85.8 / 85.9	 & \textbf{97.6 / 97.6} & 97.1 / 97.2 & 76.5 / 76.9 \\ 
    \hspace{3mm} w/ Domain adaptive & \textbf{72.1 / 72.2} &	\textbf{65.6 / 65.7} & $-$ & \textbf{56.7 / 57.1} &	65.2 / 70.4 & $-$ & \textbf{87.3 / 87.4}	& 97.5 / 97.5 & 96.9 / 97.0 & 74.9 / 75.7 \\
    
    \hline
\end{tabular}
}
\caption{\small{
Results are reported for eight different tasks, namely, Sentiment Classification (HIN-ENG\textsubscript{1} \citep{patwa2020sentimix}, HIN-ENG\textsubscript{2} \cite{patra2018sentiment}, SPA-ENG\cite{aguilar-etal-2020-lince}), Aggression Identification \cite{kumar-etal-2018-benchmarking}, Hate Speech Identification \cite{bohra2018dataset}, Offensiveness Identification \cite{chakravarthi-etal-2020-corpus}, Youtube Comments Classification \cite{kaur2019cooking}, Language Identification \cite{aguilar-etal-2020-lince}, Named Entity Recognition and Parts of Speech Tagging \cite{khanuja-etal-2020-gluecos}. ${\dagger}$ Implies results on \textit{dev} split, otherwise on \textit{test} splits.
}
}

\label{tab:benchmarking}
\end{table*}

\begin{table}[ht]
\small
\resizebox{0.48\textwidth}{!}{%
\begin{tabular}{l|rr}

\hline 

\multicolumn{1}{c|}{} & 
\multicolumn{2}{c}{\makecell{Sentiment \\ Classification}} \\ 
\cline{2-3} 
\multicolumn{1}{c|}{Model} 
& \multicolumn{1}{c}{HIN-ENG\textsubscript{1}}
& \multicolumn{1}{c}{HIN-ENG\textsubscript{2}} \\
\hline


\texttt{XLM-RoBERTa} & 68.9 / 69.1 &	61.5 / 61.5  \\ 
\hspace{3mm} w/ multi-view integration
& \textbf{71.1 / 71.3} &	62.0 / 62.8 \\ 
\hspace{3mm} w/ language-tag informed
& 68.9 / 69.3 &	62.8 / 63.1 \\ 
\hspace{3mm} w/ fasttext-BiLSTM fusion
& 69.9 / 70.0 &	61.2 / 62.1  \\ 
\hspace{3mm} w/ char-BiLSTM fusion
& 69.3 / 69.1 &	62.0 / 62.3 \\ 
\hspace{3mm} w/ semi-char-BiLSTM fusion
& 69.4 / 68.9 &	60.0 / 60.8 \\ 
\hspace{3mm} w/ data noising
& 70.5 / 70.5 &	61.9 / 62.2  \\ 
\hspace{3mm} w/ monolingual corpora
& 68.9 / 69.3 &	\textbf{68.2 / 68.3} \\ 

\hline
\end{tabular}
}
\caption{
\small{Results of various modelling techniques (F1 / Accuracy) when used with a pretrained transformers-based encoder. HIN-ENG\textsubscript{1} refers \cite{patwa2020sentimix} and HIN-ENG\textsubscript{2} refers to \cite{patra2018sentiment}}
\label{tab:all_methods}
}
\end{table}

\subsection{Input Embeddings}
\label{ssec:inputs}

\noindent \textbf{Multi-view Integration: } Tokens in mixed texts are often manifested in cross-script and mixed forms, that we refer to as \textit{views}. This infusion motivates integration of text representations in varied forms,
such as transliterated, translated, script-normalized, and
tokens belonging to one of the participating languages.
Especially in the context of pretrained multilingual models,
this technique means extracting a holistic representation
of a mixed text.
To this end, the toolkit facilitates combining representations from different \textit{views} of an input. 

\noindent \textbf{Text Tokenization: }
Motivated by some recent related works on using 
different word-level and sub-word-level
embeddings~\cite{winata2019hierarchical, aguilar-solorio-2020-english},
our toolkit offers different tokenization 
methods for encoding text.
Among the encoders available in our toolkit (\cref{ssec:models}),
pretrained transformer-based encoders 
can either be tokenized using their default 
tokenization technique (i.e. subwords) or by 
using a character-CNN architecture~\cite{boukkouri2020characterbert}.
LSTM-based models can take inputs 
in the form of tensor representations--
eg. word-level FastText~\cite{bojanowski2017enriching} 
or semi-character~\cite{SakaguchiDPD17} representations, 
or character-level representations-- 
eg. char-BiLSTM\footnote{\href{https://guillaumegenthial.github.io/sequence-tagging-with-tensorflow.html}{Sequence Tagging with Tensorflow}}.

\noindent \textbf{Tag-Informed Modeling: }
Studies in the past have shown the usefulness of 
language tag-aware modeling for mixed 
and cross-lingual texts
~\cite{chandu-etal-2018-language, lample2019crosslingual}.
However, their usefulness in the context of pretrained models and code-mixing is not thoroughly investigated.
To this end, we offer a more generalized method in our toolkit
to conduct \textit{any} tag-aware fine-tuning, 
wherein representations for different kinds of tags can be added to the text representations. Examples of such tags include
POS tags, Language IDs, etc.

\subsection{Models}
\label{ssec:models}

\noindent \textbf{Encoders: } 
An Encoder in our toolkit can consist of a 
transformer-based or BiLSTM-based architecture.
Specifically, for the former, we utilize 
pretrained models from the
HuggingFace library~\cite{wolf-etal-2020-transformers}
and the latter is implemented 
in Pytorch~\cite{paszke2019pytorch}.

\noindent \textbf{Representation Fusion: }
Encodings from different encoders can be combined,
and if required be augmented with (non-trainable) 
representations 
before passing through an adaptor. 
To combine encodings, one can either simply concat
them or obtain a (trainable) weighted average, 
a more parameter-efficient choice than the former.
Both choices are available in our toolkit.

\noindent \textbf{Adaptors: }
An adaptor is a task-specific neural layer
and currently, BiLSTM 
and Multi-Layer Perceptron (MLP) choices are available 
as part our toolkit.
The inputs to adaptors are 
fused representations
if multiple
encoders are specified, else output from a single 
encoder.
These adaptors serve as task-specific learnable parameters.

\noindent \textbf{Multitasking: }
Multi-task learning can help models to pick
relevant cues from one task to be applied to another.
Such a setting was also previously investigated 
in the context of mixed texts,
which showed promising improvements~\cite{chandu-etal-2018-language}.
Furthermore, it is also shown in monolingual 
NLP that incorporating
explicit semantics 
as an auxiliary task
can enhance BERT's 
performance~\cite{zhang2020semanticsaware}.
Motivated by these, our toolkit offers support 
to conduct training of one or more
tasks. Once a final representation is
produced by adaptors of each task,
we use a training criterion to compute loss 
and perform gradient backpropagation.

\subsection{Tasks}
\label{ssec:tasks}

\noindent \textbf{Tasks: }
Our toolkit currently supports two kinds of tasks-- 
sentence-level text classification 
and word-level sequence tagging, the flow for each is demonstrated in Figure \ref{fig:components}.
The \textit{decoupled} design of our toolkit helps 
in seamlessly creating multi-task training setups. 
The kinds of tasks for which we offer support currently are listed in Table \ref{tab:benchmarking}.


\noindent \textbf{Adaptive Pretraining: }
Following the successes of task-adaptive
and domain-adaptive pretraining in monolingual and multilingual NLP tasks
\cite{dontstoppretraining2020},
users of our toolkit can also perform such adaptive pretrainings
using mixed texts on top of pretrained transformer-based models.

\subsection{Codemixed Quantification}
\label{ssec:quantification}

Our toolkit offers 5 standardized metrics for
quantifying mixing in text, 
namely Code-Mixing Index ~\cite{gamback2014measuring}, 
Average switch-points ~\cite{khanuja-etal-2020-gluecos}, 
Multilingual Index, 
Probability of Switching and Language Entropy ~\cite{guzman2017metrics}.
We offer simple command line methods to compute these metrics
and also offer metric-based data sampling.

\subsection{Data Augmentation}
\label{ssec:data_augmentation}

Our toolkit also offers techniques to do data augmentation. 
While data augmentation is useful in cases where there is 
training data scarcity, for mixed datasets,
it is also essential to produce a more generalized model.
As part of this feature, this toolkit currently offers
augmentation through transliteration, 
spelling variations and monolingual corpora.
We currently support transliteration of Indic languages
through an off-the-shelf tool- \texttt{indic-trans} \cite{Bhat:2014:ISS:2824864.2824872}.
Spelling variations include noising spelling,
such as randomly removing/replacing vowel characters.
Monolingual corpora augmentation is task specific.
For a given task, such as sentiment classification, we 
augment publicly available monolingual corpora 
based on the task 
type
from 
one or all of the mixing languages
and use it while fine-tuning models.

\subsection{Data Format}
\label{ssec:data_format}

Due to diverse data formats of existing mixed datasets,
benchmarking and comparing results across tasks is not readily feasible.
To this end, we propose a standardized data format for 
 syntactic, semantic level understanding and generation tasks, and our toolkit
offers  command line methods to adopt a user's 
dataset to this standard format.

\section{Experiments}
\label{sec:experiments}

Among many potential research applications of our toolkit,
in this section, we demonstrate one -- \textit{benchmarking}.
Table~\ref{tab:benchmarking}
presents 
performances of selected 
model architectures 
obtained using our toolkit
on some
popular mixed datasets.
In Table~\ref{tab:all_methods},
we also demonstrate the performances of 
different architectural choices
implemented through our toolkit
on two Hinglish datasets.
For domain-adaptive pretraining of Hinglish datasets,
we collate around 160K mixed sentences from several of the 
publicly available Hinglish datasets.
For task-adaptive pretraining, 
we just use the training and testing data available in the 
dataset of interest.
For training, we use standard optimizers and model configurations.
\footnote{Due to the space limitations, we direct the reader to 
check \href{https://github.com/murali1996/CodemixedNLP}{github.com/murali1996/CodemixedNLP}
for toolkit usage and for the list of modeling choices.}

\section{Demo}
\label{sec:demo}

\begin{figure}
\centering
\includegraphics[width=0.5\textwidth]{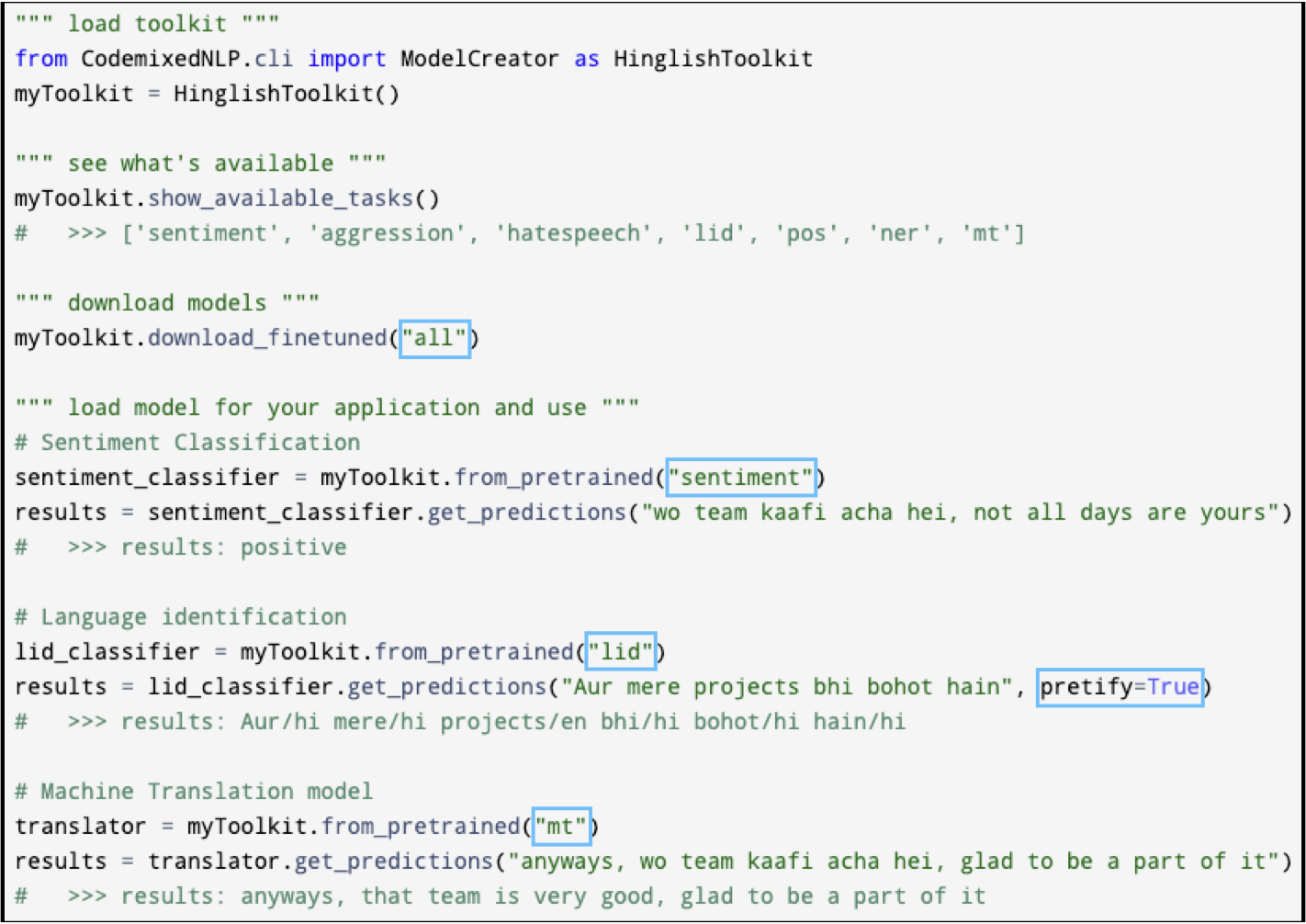}
\caption{\label{fig:toolkit_cmd} 
\small Command line interface for utilizing fine-tuned models. We provide several functionality compatible with the popular Huggingface and Fairseq libraries. Marked in boxes are customizable input arguments.
}
\end{figure}

We fine-tune and publish transformer-based models
for 7 tasks in Hinglish. We include 3 task types--
(1) \textbf{Semantic} (Sentiment Classification, Hate Speech  and Aggression Identification),
(2) \textbf{Syntactic} (NER, POS and Language Identification),
and 
(3) \textbf{Generation} (Hinglish$\rightarrow$English Machine Translation).
We present some examples of utilizing these models 
in Figure~\ref{fig:toolkit_cmd}.


\section{Conclusion}
\label{sec:conclusion}

In this work, we presented a unified toolkit for modeling code-mixed texts. 
Additionally, the toolkit contains various functionalities such as data augmentation, 
code-mixing quantification, and ready-to-use fine-tuned models for
7 different NLP tasks in Hinglish.
Our toolkit is simple enough for practitioners to integrate new features as well as develop wrappers around its existing functionalities. We believe this contribution facilitates a sustainable and extensible ecosystem of models by adding novel pretraining techniques tailored for mixed texts,
text normalization techniques to counter spelling variations, error analysis tools to identify peculiarities in incorrect predictions and so on.



\bibliography{anthology,emnlp2020}
\bibliographystyle{acl_natbib}

\end{document}